\documentclass[10pt,twocolumn,letterpaper]{article}

\usepackage{cvpr}              % To produce the CAMERA-READY version
\usepackage{graphicx}
\usepackage{amsmath}
\usepackage{amssymb}
\usepackage{booktabs}

\usepackage{color}
\usepackage{comment}
\usepackage{multirow}
\usepackage{amsmath}
\usepackage{bbm}

\newcommand{\blue}[1]{{\color{blue} #1}}

\usepackage[pagebackref,breaklinks,colorlinks]{hyperref}

\usepackage[capitalize]{cleveref}
\crefname{section}{Sec.}{Secs.}
\Crefname{section}{Section}{Sections}
\Crefname{table}{Table}{Tables}
\crefname{table}{Tab.}{Tabs.}

\def\cvprPaperID{***} % *** Enter the CVPR Paper ID here
\def\confName{CVPR}

\begin{document}

\title{\texttt{\textbf{CAVL}}: Learning \texttt{\textbf{C}}ontrastive and \texttt{\textbf{A}}daptive Representations of \\ \texttt{\textbf{V}}ision and \texttt{\textbf{L}}anguage}

\author{%
  Shentong Mo\thanks{These authors contributed equally to this work.}, Jingfei Xia$^*$, Ihor Markevych \\
  Carnegie Mellon University
}

\maketitle

\begin{abstract}

% background
Visual and linguistic pre-training aims to learn vision and language representations together, which can be transferred to visual-linguistic downstream tasks.
% motivation
However, there exists semantic confusion between language and vision during the pre-training stage.
Moreover, current pre-trained models tend to take lots of computation resources for fine-tuning when transferred to downstream tasks.
In this work, we present a simple but effective approach for learning \texttt{\textbf{C}}ontrastive and \texttt{\textbf{A}}daptive representations of \texttt{\textbf{V}}ision and \texttt{\textbf{L}}anguage, namely \texttt{\textbf{CAVL}}.
Specifically, we introduce a pair-wise contrastive loss to learn alignments between the whole sentence and each image in the same batch during the pre-training process.
At the fine-tuning stage, we introduce two lightweight adaptation networks to reduce model parameters and increase training speed for saving computation resources.
We evaluate our CAVL on six main downstream tasks, including Visual Question Answering (VQA), Visual Commonsense Reasoning (VCR), Natural Language for Visual Reasoning (NLVR), Region-to-Phrase Grounding (RPG), Text-to-Image Retrieval (TIR), and Zero-shot Text-to-Image Retrieval (ZS-TIR).
Compared to baselines, we achieve superior performance and reduce the fine-tuning time by a large margin (in particular, 76.17\%).
Extensive experiments and ablation studies demonstrate the efficiency of contrastive pre-training and adaptive fine-tuning proposed in our CAVL.

\end{abstract}

\section{Introduction}
% \vspace{-0.5em}
Visual and language representations pre-training~\cite{li2019visual,lu2019vilbert,Su2020VL-BERT:} has been an active research area in the multi-modal community. 
This is because it allows for the usage of pre-trained models that achieve state-of-the-art comparable results for a variety of tasks without spending significant compute time for modeling language and visual distributions by leveraging features created by available pre-trained models.
With a better pre-training model, it can be used in a variety of areas in visual and language fields such as Visual Question Answering (VQA)~\cite{goyal2017vqa} and Visual Commonsense Reasoning (VCR)~\cite{zellers2019vcr}. 
In various architectures designed for different visual-linguistic tasks, a key point is to aggregate the multi-modal information in both visual and linguistic domains. 
However, there exists \textit{semantic confusion} between vision and language at the pre-training stage, that is, misalignment between object/entities or image/text. 
Another problem is that, when transferred to downstream tasks, pre-trained models tend to take much training time and resources for fine-tuning. 

\begin{figure}
\setlength{\abovecaptionskip}{0em}
\setlength{\belowcaptionskip}{-1em}
	    \centering
% 		\fbox{\rule{0pt}{2in} \rule{0.8\linewidth}{0pt}}
	\includegraphics[width=0.9\linewidth]{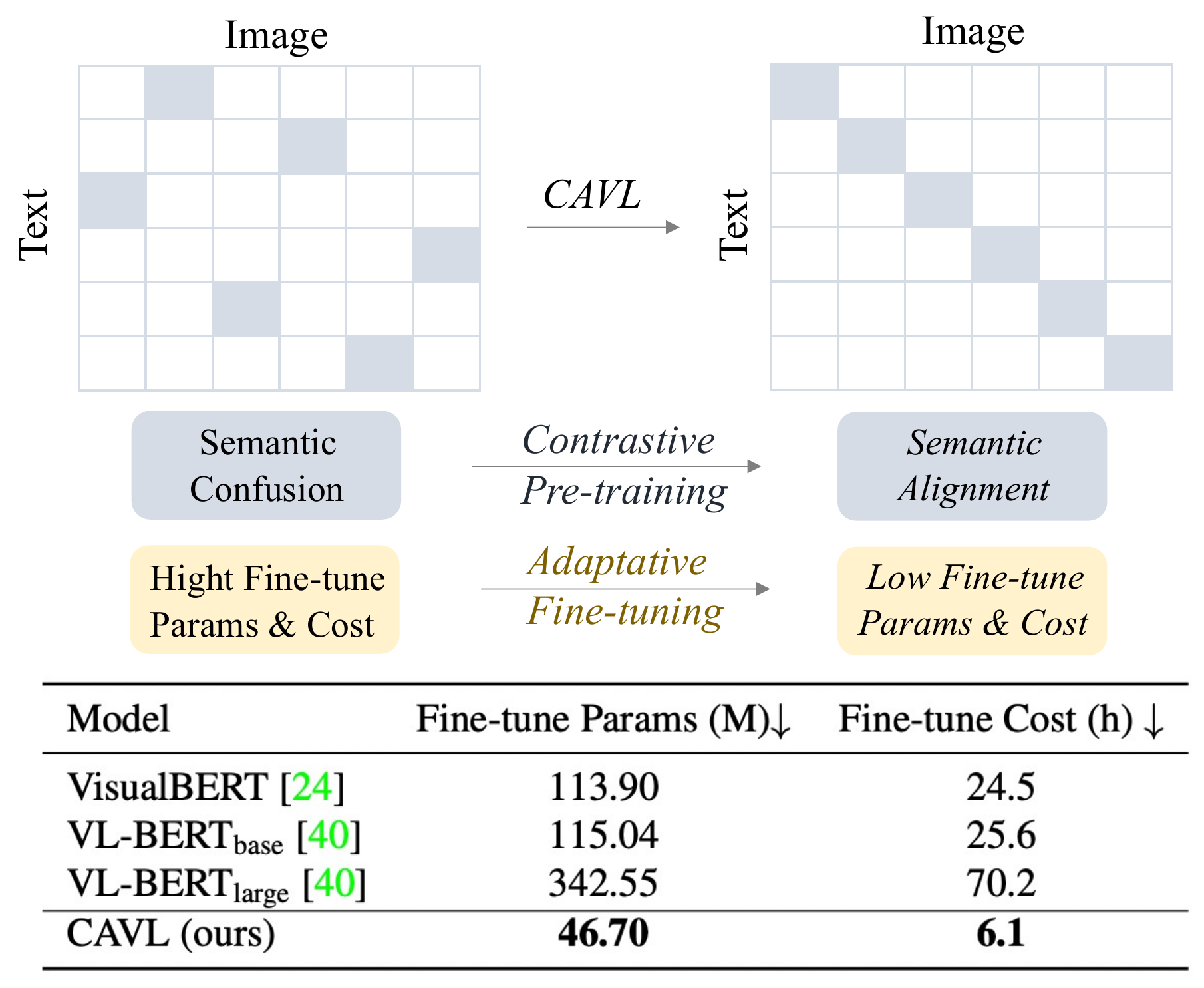}
  \caption{Illustration of two main problems (semantic confusion and high fine-tuning cost) existing in vision-language models, where we propose the \texttt{\textbf{CAVL}} (contrastive pre-training and  adaptative fine-tuning) to address the issues. }
	\label{fig: title_img}
\end{figure}

% contrastive learning motivation
In this work, we propose a simple but effective framework for learning \texttt{\textbf{C}}ontrastive and \texttt{\textbf{A}}daptive representations of \texttt{\textbf{V}}ision and \texttt{\textbf{L}}anguage, called \texttt{\textbf{CAVL}}, which involves contrastive pre-training and adaptive fine-tuning. 
Specifically, to eliminate the misalignment between language and vision during pre-training, we apply a Pair-wise Contrastive Loss (PwCL) to learn alignments between the whole sentence and each image, where we maximize the cosine similarity of visual and linguistic embeddings from correct pairs while minimizing the cosine similarity of embeddings of false pairs.
To further eliminate the need for much training time at the fine-tuning stage, we introduce two lightweight adaptation networks in our CAVL to learn adaptive representations. 
One adapter is to use a shortcut block to obtain task-specific features and merge generalized features from the pre-trained model with the output block; the other adapter is to apply a bottleneck structure after attention and feed-forward modules in each layer in BERT. 
Our CAVL not only reduces the training parameters greatly but also maintain the performance at a competitive level. 

We conduct extensive experiments on four main downstream tasks: VQA, VCR, Natural Language for Visual Reasoning, and Region-to-Phrase Grounding.
Compared to previous state-of-the-art models~\cite{li2019visual,lu2019vilbert,Su2020VL-BERT:,tan2019lxmert,uniter,shi2020contrastive}, our CAVL achieves comparable or even better performance when transferred to visual-linguistic downstream tasks. 
Contrastive pre-training assists our CAVL in better understanding the relationship between image and text which improves results on downstream tasks.
Ablation studies on adaptive fine-tuning also demonstrate the effectiveness and efficiency of the proposed adaptive fine-tuning in saving computation resources. 

Overall, our main contributions in this work can be summarized as follows:
\begin{itemize}
    \item We propose a simple but effective approach for learning alignments between visual and linguistic representations during pre-training, namely CAVL. 
    \item We present two lightweight adaptation networks in our CAVL to further ease the need for large computation resources at the fine-tuning stage.
    \item Our CAVL achieves superior results when transferred to six main visual-linguistic downstream tasks.
    \item  Extensive ablation studies demonstrate the efficiency of adaptive fine-tuning in reducing training parameters while achieving comparable performance.
\end{itemize}

\section{Related Work}

\noindent{\textbf{Visual representations pre-training.}}
In recent years, visual representations pre-training has been applied to many downstream tasks, such as image classification, object detection, and segmentation. 
Typically, contrastive self-supervised learning~\cite{chen2020simple,chen2020big,chen2020mocov2,khosla2020sup,mo2021spcl,mo2022pauc,mo2023mcvt} is one of the popular methods to learn meaningful visual representations.
Most previous methods learn visual representations from text paired with images in unsupervised, self-supervised, weakly supervised, and supervised.
Since language and vision can share a similar semantic meaning, CLIP~\cite{radford2021learning} is a commonly-used neural network trained on a variety of (image, text) pairs for learning transferable visual representations from natural language supervision. 
With the instruction of natural language and task-agnostic optimization, CLIP can predict the most relevant text snippet given an image, which is similar to the zero-shot capabilities of GPT-2~\cite{radford2019gpt2} and GPT-3~\cite{brown2020language}.
Huo \textit{et al.}~\cite{huo2021wenlan} apply a cross-modal contrastive learning framework called BriVL for image-text pre-training. Unlike CLIP that adopts a simple contrastive learning method, they make use of MoCo~\cite{he2019moco} for the cross-modal scenario by building a large queue-based dictionary to introduce negative samples in a batch for the pre-training.
In this work, misalignments between visual and linguistic embeddings are mitigated by a pair-wise contrastive loss at the pre-training stage for the multi-modal scenario.

\noindent{\textbf{Linguistic representations pre-training.}}
In the language modeling literature, there are two main frameworks for linguistic representations pre-training, including BERT~\cite{devlin2018bert} and GPT~\cite{radford2018gpt}.
BERT~\cite{devlin2018bert} is a transformers-based~\cite{ashish2017attention} pre-trained model, which advanced state-of-the-art results for a lot of natural language processing tasks with self-attention module.
GPT~\cite{radford2018gpt} is another language modeling architecture that is based on transformers~\cite{ashish2017attention}. 
Input for GPT models is represented on a byte level with an exception of spaces which allows handling variable vocabularies and deal with unknown during training time tokens. 
In this work, we introduce a simple but effective framework based on BERT and evaluate our network on four main visual-linguistic downstream tasks. 

\noindent{\textbf{Fusion of visual and linguistic representations pre-training.}}
There is a bunch of work~\cite{imagebert,kvlbert,ernievil,uniter,sun2019videobert,tan2019lxmert,li2019visual,lu2019vilbert,Su2020VL-BERT:,hong2021recurrent,lei2021less,li2020oscar,Yu2021ERNIEViL,zhang2021vinvl,Jia2021ALiGN} focusing on the fusion of visual and linguistic representations pre-training.
Typically, UNITER~\cite{uniter} aims at learning to join image-text embeddings by using transformer on multi-model inputs with Masked Language Modeling (MLM), Masked Region Modeling, Image-Text Matching, and Word-Region Alignment as pre-training tasks.
ERNIE-ViL~\cite{Yu2021ERNIEViL} uses the joint distributions of vision-language with scene graphs of visual tasks to predict nodes of different types of scene graph parsed from the sentence.
LXMERT~\cite{tan2019lxmert} is a transformer based model with encoders: an object relationship encoder, a language encoder, and a cross-modality encoder.
VisualBERT~\cite{li2019visual} is a simple and flexible framework used for vision-and-language tasks, where they use the self-attention module in BERT structure to combine the image embedding in vision and text embedding in language. ViLBERT~\cite{lu2019vilbert} extends the traditional BERT by using two parallel BERT-type models that operate over text segments and image regions. 
VL-BERT~\cite{Su2020VL-BERT:} is another BERT-based model that takes regions of interest from images and sub-word information, where they pre-trains the model by predicting masked words with image clues and predicting masked regions with text clues.
To address the noisy label and domain bias problems, CVLP~\cite{shi2020contrastive} introduces a contrastive loss in the visual branch to discriminate between positive examples and negative ones.
More recently, DocFormer~\cite{srikar2021docformer} is proposed to enforce the multi-modal interaction between visual, text and spatial features for Visual Document Understanding.
However, we adopt the pair-wise contrastive loss in both visual and linguistic branches to eliminate the misalignment between the whole sentence and each image during pre-training. 

Note that our CAVL is different from a recent work, ConVIRT~\cite{zhang2020contrastive}, in two major ways.
1) Similar to CLIP, ConVIRT applies the contrastive loss to train a model that performs better in medical image classification with natural language supervision while in this work we focus on solving the misalignment problems in visual and linguistic area. 2) They generate visual and textual embeddings via an image encoder and a text encoder, separately.
However, we combine two encoders into single BERT model and show that the pair-wise contrastive learning enables model to learn better aligned representations of image and text. 
Moreover, our PwCL loss differs from the VLM loss proposed in Unicoder-VL~\cite{li2020unicodervl} in two ways.
1) Unicoder-VL introduces FC layers to predict the score between the whole image and sentence, while our PwCL calculates the cosine similarity by a simple dot-product without any additional parameters.
2) The VLM loss just samples one negative sample (image or caption) and applies a cross-entropy loss as binary classification to learn the alignments.
However, we have $B^2 - B$ negative samples in the whole batch of size $B$ during pre-training such that the misalignment problems are more fully addressed.

\noindent{\textbf{Adaptation networks in transformers.}}
Adaptation is an important way when fine-tuning the BERT model, which allows us to achieve a promising result by updating much fewer parameters with less time and computation resources.
This motivates researchers to use efficient adaptation methods in transformer-based models. 
MAD-X~\cite{pfeiffer2020madx} adapts a multilingual model to arbitrary tasks and languages, where authors propose an invertible adapter and show good performance on different languages and tasks.
Houlsby \textit{et al.}~\cite{houlsby2019parameterefficient} propose an intermediate layer inside transformer layers and train the intermediate layer with all other parameters frozen. 
In this work, we present two lightweight adapters to achieve competitive performance on domain-specific tasks with significantly reduced fine-tuning parameters and costs.

\section{Method}

% \vspace{-0.5em}

\subsection{Preliminary: BERT}

BERT~\cite{devlin2018bert}, which stands for Bidirectional Encoder Representations from Transformers, is a model that uses word tokens as input and optimizes language modeling objectives. 
All input word tokens are mapped to a set of embeddings and passed through several ``encoder-style'' blocks to generate representations. 
The input embeddings are calculated as the sum of token embedding, segment embedding, and position embedding. 
Then the input embeddings are passed through a multi-layer transformer, where each layer consists of two modules: multi-head attention and feed-forward module. 
Each module is followed by a fully connected network and together wrapped in a residual addition.

There are two main steps for the BERT training process: pre-training and fine-tuning. 
Two language modeling objectives are used for the pre-training step: (1) Masked Language Modeling (MLM). 
The embedding of a certain input word would be randomly masked out 
and BERT is used to predict the masked words using the unmasked words.
(2) Next Sentence Prediction (NSP). 
BERT is learned to identify whether two sentences are consecutive sentences from one document given two sentences. 
Finally, in order to apply BERT to a particular task, we introduce a task-specific input, an output layer, an objective, and the fine-tuned model with a specific dataset.

% \vspace{-0.5em}
\subsection{\texttt{\textbf{CAVL}}}
% \vspace{-0.5em}
In this part, we present a simple but effective approach for learning contrastive and adaptive representations of vision and language, namely CAVL, which consists of contrastive pre-training and adaptive fine-tuning.
Specifically, contrastive pre-training is applied to mitigate the semantic confusion between visual and linguistic representations at the pre-training stage.
When transferred to downstream tasks during the fine-tuning process, adaptive fine-tuning eliminates the need for much training time and large GPU memories. 

% \vspace{-1em}

\subsubsection{Contrastive Pre-training}

\begin{figure*}[!htb]
\setlength{\belowcaptionskip}{-0.5em}
    \centering
    \includegraphics[width=0.9\textwidth]{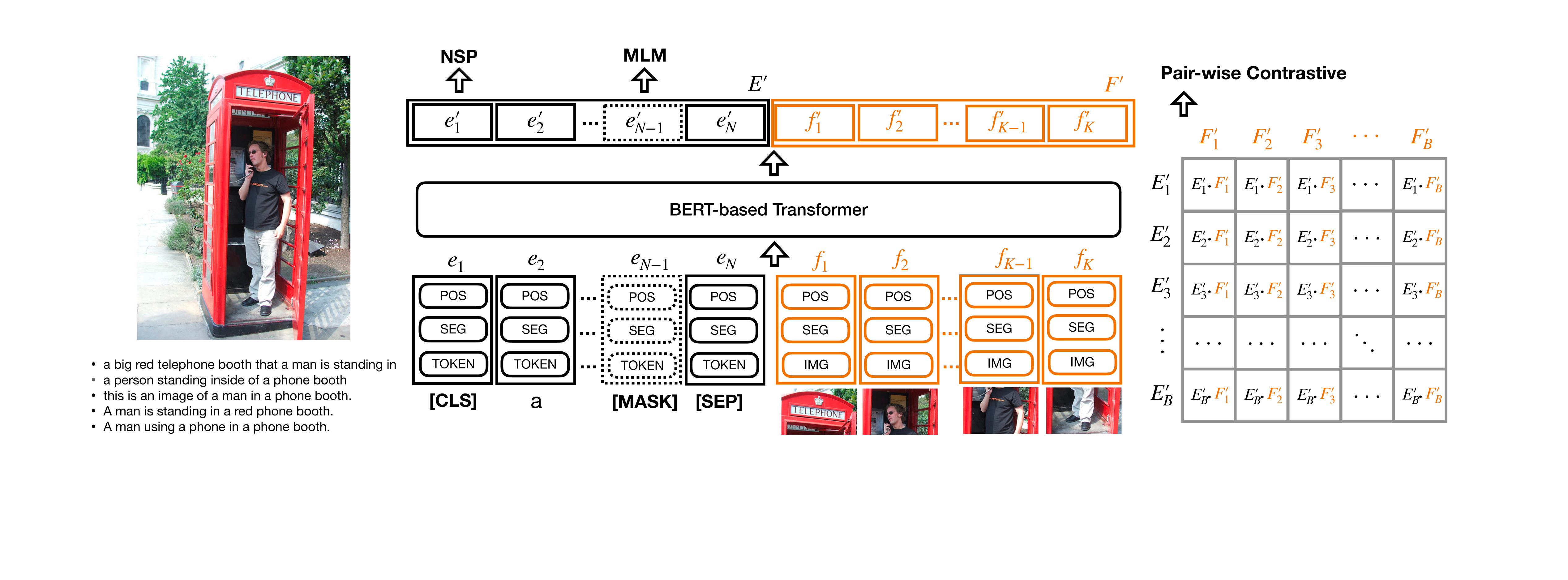}
    \caption{The overview of contrastive pre-training  of our CAVL for pair-wise visual-linguistic representations. A self-attention mechanism within the BERT-based Transformer is applied to implicitly align elements of the input text and regions in the input image, and a contrastive learning framework is used for alignments between the whole sentence and each image in an explicit manner.}
    \label{fig:overview}
\end{figure*}

% \vspace{-0.5em}
For contrastive pre-training, we adopt a self-attention mechanism within the BERT-based transformer to explicitly
align elements of the input text and regions in the input image in a contrastive manner, as can be illustrated in Figure~\blue{\ref{fig:overview}}.
Our CAVL consists of three main components: linguistic pre-training, visual pre-training, and contrastive fusion of visual-linguistic pre-training. 

\textbf{Linguistic pre-training.}
For language embeddings in the pre-training, we input three types of embeddings: 
1) a token embedding $\mathbf{e}^t$ for each subword in a sentence; 
2) a segment embedding $\mathbf{e}^s$ indicating which part of the text the token is from; 
3) a position embedding $\mathbf{e}^p$ for the position of the token in the sentence. 
Then we sum up all three embeddings in a contextual representation $\mathbf{e}_{n}, n \in\{1,2,...,N\}$, where $N$ denotes the number of subwords in the sentence.
After being fed into a BERT-based transformer, those contextual embeddings become $\mathbf{e}^\prime_{i}$.
We adopt two similar objectives as BERT, including masked language modeling (MLM) and next sentence prediction (NSP).
For the former objective, we randomly mask some
parts of the input tokens with a special token (\textit{i.e.}, [MASK]), and the model is trained to predict the masked token.
As for NSP, we train the model using the embedding [CLS] to classify whether a pair of given sentences are two consecutive sentences in a context.

\textbf{Visual pre-training.}
For vision features in the pre-training, we extract image ROIs from an objection detection framework (\textit{i.e.}, Faster R-CNN) as the input $\mathbf{f}_k, k\in \{1, 2, ..., K\}$, where $K$ is the number of image ROIs. 
The input $\mathbf{f}_k$ is also composed of three types of visual embeddings:
1) an image feature embedding $\mathbf{f}^i$ for each image ROI;
2) a segment embedding $\mathbf{f}^s$ indicating which token embedding the image embedding is opposed to; 
3) a position embedding $\mathbf{f}^p$ for alignments between tokens and each image ROI.
Following Visual-BERT in task-agnostic pre-training, we apply two captions for a text segment in the COCO dataset, where there are multiple captions corresponding to one image, as can be seen in Figure \blue{\ref{fig:overview}}. 
Particularly, we use one of the captions as ground truth to describe the image, while we also apply a 50\% probability to choose a caption from those two captions.
Our model is trained to distinguish whether the caption is the ground truth or randomly drawn from two captions.

\textbf{Visual-linguistic pre-training.}
To mitigate the semantic confusion between language and vision, we design a pair-wise contrastive learning mechanism on visual and linguistic representations from the multi-layer transformer.
Specifically, we calculate the cosine similarity between each pair of linguistic embeddings $\mathbf{E}_b^\prime$ and visual embeddings $\mathbf{F}_b^\prime$ in a batch of size $B$, where $b\in \{1, 2, ..., B\}$.
Then, those similarities are jointly learned for alignments between the whole sentence and each image in the same batch, where we maximize the cosine similarity of the visual and linguistic embeddings of the $B$ correct pairs in the batch while minimizing the cosine similarity of the embeddings of the $B^2-B$ false pairings. 
We apply a pair-wise contrastive loss over these similarities scores for optimization.

Specifically, we define the Pair-wise Contrastive Loss (PwCL) between linguistic embeddings $\mathbf{E}_i^\prime$ and visual embeddings $\mathbf{F}_j^\prime$ as: 
\begin{equation}
\begin{aligned}
\mathcal{L}_{\text{PwCL}} = -\log \dfrac{\sum_{i=1}^B (\mathbf{E}_i^\prime \cdot \mathbf{F}_i^\prime)}{\sum_{i=1}^B\sum_{j=1}^B \mathbbm{1}_{i\neq j}(\mathbf{E}_i^\prime \cdot \mathbf{F}_j^\prime)}
\end{aligned}
\end{equation}
where $\mathbbm{1}_{i\neq j}$ is an indicator function to check if linguistic embeddings $\mathbf{E}_i^\prime$ and visual embeddings $\mathbf{F}_j^\prime$ are aligned or not. 
In this way, we maximize the cosine similarity of visual and linguistic embeddings from correct pairs while minimizing the cosine similarity of embeddings of false pairs.
Intuitively, alignments between the whole sentence and each image are learned in our CAVL to mitigate the semantic confusion existing in the pre-training process. 

It is worthy mentioning that this kind of contrastive pre-training methodology is different from a recent powerful framework, CLIP~\cite{radford2021learning}. 
Concretely, we build a image-text alignment pre-training framework for the multi-modal scenario.
First, this idea in CLIP is used to train a model that performs better in image classification with natural language supervision while in this paper we focus on solving multi-modal problems in visual and linguistic area. 
Second, they train an image encoder and a text encoder to generate vision and language embeddings. 
However, we combine two encoders into one BERT model and show that pair-wise contrastive learning enables model to learn better representations of image and text.

\begin{figure*}[!htb]
\setlength{\belowcaptionskip}{-0.5em}
    \centering
    \includegraphics[width=0.5\textwidth]{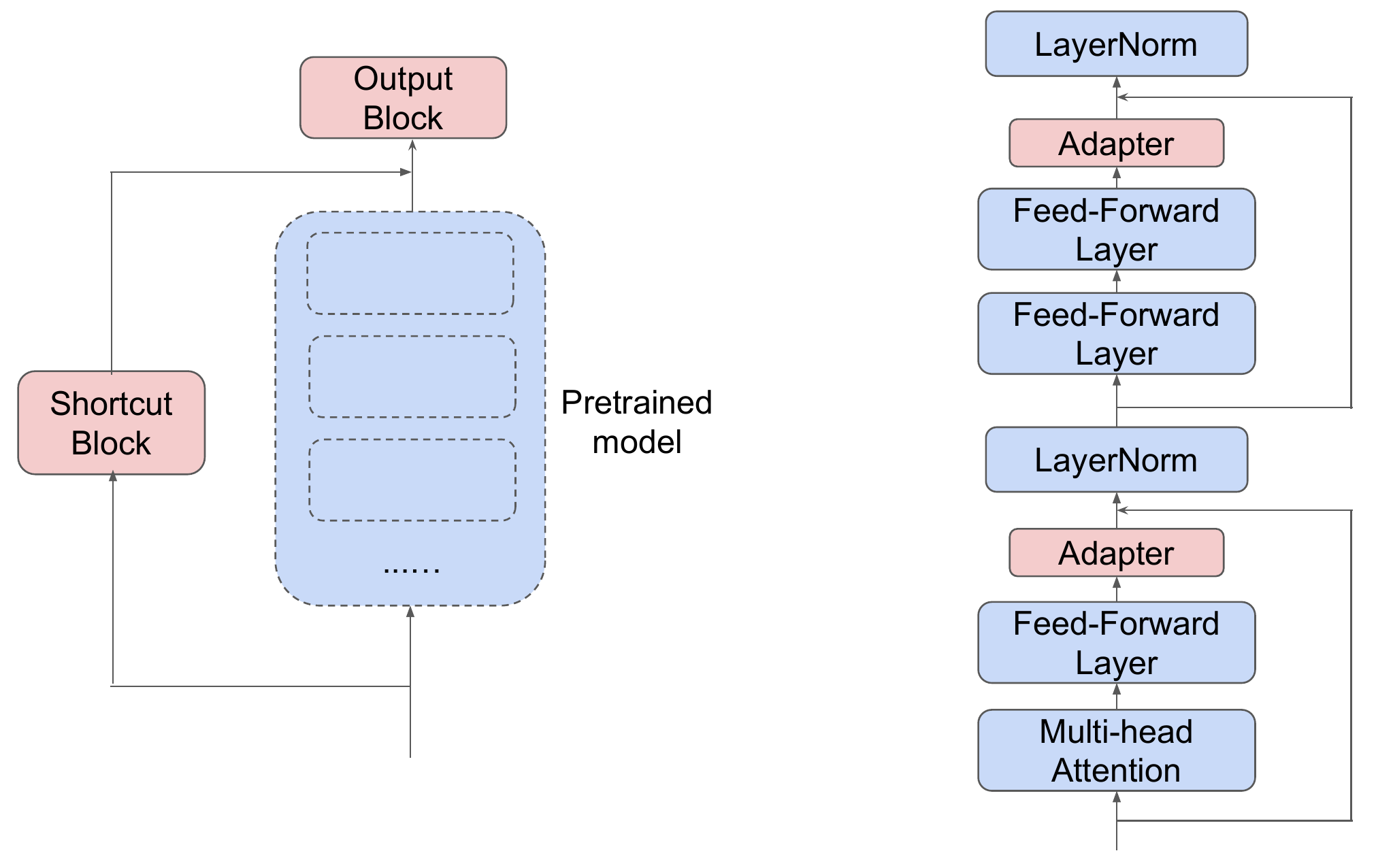}
    \caption{Adaptive Fine-tuning with Adapters in CAVL. Adapter I (\textbf{Left}): This adaptation method freezes the pre-trained model and trains a shortcut block and output block during fine-tuning. Adapter II (\textbf{Right}): This adaptation method follows \cite{houlsby2019parameterefficient} which adds an adapter right before each LayerNorm layer in the BERT. During the fine-tuning, it freezes feed-forward layers and attention modules while updating parameters in the adapter and LayerNorm layer. Blue blocks denote the frozen part and parameters in red blocks are updated during the fine-tuning phrase.}
    \label{fig:adapter}
\end{figure*}

\subsubsection{Adaptive Fine-tuning}

In this section, we design one adaptation methods and compared that with other efficient methods to fine tune on down-streaming tasks. 
One idea is that the frozen part in the model provides basic information for the generalized problem while the updated part generates a task-specific feature. 
Inspired by this idea, we proposed a method that used a shortcut block aside from the pre-trained model and merges the output from the pre-trained model and shortcut block with an additional output block. 
The pre-trained model obtains generalized features between image and text, and the shortcut block acts as the selection neuron to capture the feature in each specific task. 
Then we apply an output block to combine generalized feature and task-specific feature together to get the final result. 
There are fewer layers in shortcut and output block than those in pre-trained model so the training time is reduced. 
We denote this type of adapter as \textbf{Adapter I}. 

% \vspace{-1em}

Motivated by Neil \textit{et al.}~\cite{houlsby2019parameterefficient}, we propose another adaptation method denoted as \textbf{Adapter II} by adding a bottleneck structure within each BERT layer. 
Specifically, an adapter is added at the end of the attention module and another one at the end of the feed-forward layer. 
The adapter output and input of both the attention module and feed-forward layer are added together to pass through the LayerNorm module. 
During the fine-tuning process, we freeze the attention module and feed-forward layer. 
So, the adapter acts as a projection module which reflects the generalized feature to task-specific features.
The adapter with fewer parameters is designed as a bottleneck structure, which contains one linear layer, a GELU function, and another linear layer. 

We show the network details about two types of adapters for adaptive fine-tuning in Figure \blue{\ref{fig:adapter}}. We also conduct a comprehensive ablation study on the adapters in Section \blue{\ref{ab:adpater}}.

\begin{table*}[t]
    \centering

	\renewcommand\tabcolsep{6.0pt}
	\scalebox{0.8}{
	\begin{tabular}{lllllllll}
		\toprule
		\multicolumn{1}{l}{\multirow{3}{*}{Model}} & \multicolumn{2}{c}{VQA}  & \multicolumn{6}{c}{VCR} \\
		& test-dev & test-std & \multicolumn{2}{c}{Q $\rightarrow$ A} & \multicolumn{2}{c}{QA $\rightarrow$ R} & \multicolumn{2}{c}{Q $\rightarrow$ AR}\\
		& & & val & test & val & test &  val & test \\ 
		
		\midrule
		
		LXMERT~\cite{tan2019lxmert} & 72.42 & 72.54 & - & - &- & - & - & -\\
		ViLBERT~\cite{lu2019vilbert}  &70.55 & 70.92 & 72.40 & 73.30 & 74.50 & 74.60 & 54.00 & 54.80 \\
		VisualBERT~\cite{li2019visual} &70.08 & 71.00 & 70.80 & 71.60 & 73.20 & 73.20 & 52.20 & 52.40 \\   
        VL-BERT~\cite{Su2020VL-BERT:}  &71.72 & 72.18 & 73.80 & - & 74.40 & -& 55.20 & -\\
		UNITER~\cite{uniter} & 72.27 & 72.46 & 	-& 75.00	& - & 77.20 & - & 58.20\\
		CVLP~\cite{shi2020contrastive} & 72.77 & 72.90 & - & - & - & - & - & -\\
		CAVL (ours)  & \textbf{72.83} & \textbf{73.05} & \textbf{75.33} & \textbf{75.65} & \textbf{76.52} & \textbf{77.87} & \textbf{58.65} & \textbf{59.47}\\
		\bottomrule
		\end{tabular}
	}
	\caption{Comparison results on the VQA and VCR datasets.}
	\label{tab: exp_vqa_vcr}
	\vspace{-0.5em}
\end{table*}

\section{Experiments}
\subsection{Pre-training \& Implementation Details}

Following previous work~\cite{li2019visual,lu2019vilbert,Su2020VL-BERT:}, we apply the same setting for a fair comparison with those baselines. 
Specifically, we pre-train our CAVL on MS COCO~\cite{coco} and Visual Genome~\cite{krishna2017visualgenome}.
For visual tokens, we apply the pre-trained Faster R-CNN~\cite{ren2015faster} to extract the image ROIs (at most 100 ROIs with detection scores higher than 0.5 for each image).
We apply SGD with Adam~\cite{kingma2014adam} for optimization and use a total batch size of 512 for 10 epochs. 
We use the warm-up step number of 15\% of the total training steps.
The pre-training and fine-tuning costs are 88 and 10 hours on 4 Tesla V100-32G GPUs respectively.

\subsection{Downstream Tasks}
We evaluate our CAVL pre-trained models on four downstream tasks: (1) Visual Question Answering (VQA), (2) Visual Commonsense Reasoning (VCR), (3) Natural Language for Visual Reasoning (NLVR$^2$), and (4) Region-to-Phrase Grounding (Flickr30K).
Unless otherwise specified, we adopt Adapter I as the adaptive fine-tuning method.

\noindent{\textbf{Visual Question Answering (VQA).}}
In the VQA task, we follow the experimental protocol
in BUTD~\cite{anderson2018butd},
aim to answer a question at the perceptual level given a natural image by choosing the correct answer from a shared set composed of 3,129 answers. 
Specifically, we conduct experiments on the VQA
v2.0 dataset~\cite{goyal2017vqa} based on the images from COCO~\cite{coco} dataset. 
We split the dataset into train set (83k images and 444k questions), validation set (41k images and 214k questions), and test set (81k images and 448k questions). 
We report the results in Table~\ref{tab: exp_vqa_vcr}.
Compared to previous methods, our CAVL achieves better performance in terms of accuracy on both test-dev and test-std datasets. 
This infers that the pair-wise contrastive loss between visual and linguistic representations is beneficial for learning alignments between the whole sentence and each image.

% \vspace{-0.5em}
\noindent{\textbf{Visual Commonsense Reasoning (VCR).}}
In the VCR task, we need to select the right answer to the given question and provide the rationale explanation for a higher-level cognitive and commonsense understanding of the given image. 
In the experiments, we use an image and a list of categorized ROIs from the VCR dataset~\cite{zellers2019vcr} to pick the correct one from 4 candidate answers and 4 candidate rationales, respectively.
The task (Q $\rightarrow$ AR) can be split into two sub-tasks: question answering (Q$\rightarrow $A) and answer justification (QA$\rightarrow$R).
We also split the VCR dataset into training (213k questions and 80k images), validation (27k questions and 10k images), and test (25k
questions and 10k images) sets.
The results are reported in Table~\ref{tab: exp_vqa_vcr}.
Our CAVL achieves competitive performance, although we do not use the larger Conceptual Captions dataset in VL-BERT$_\text{large}$. 
This implies that the pair-wise contrastive loss proposed at the pre-training stage is beneficial to eliminate the semantic confusion between vision and language.
VL-BERT$_\text{large}$ also validates the importance of pre-training on a massive-scale dataset to improve the model's capacity.

\begin{table}[t]
	%\normalem
	\renewcommand\tabcolsep{6.0pt}
	\centering
	\scalebox{0.8}{
	\begin{tabular}{lllll}
		\toprule
		Model & Dev & Test-P & Test-U & Test-U (Cons)  \\ 	\midrule
	
		LXMERT~\cite{tan2019lxmert} & - &74.45 & 76.20 & 42.10\\
		VisualBERT~\cite{li2019visual}  &67.40 &67.00 &67.30 &26.90 \\
		 UNITER~\cite{uniter} & 76.93 & 75.58 & - & - \\
		 CVLP~\cite{shi2020contrastive} & - & 76.20 & - & -\\
		CAVL (ours)  & \textbf{79.16} & \textbf{78.31} & \textbf{79.87} & \textbf{46.23}\\
		\bottomrule
		\end{tabular}
	}
	\caption{Comparison results on the NLVR$^2$ dataset.}
	\label{tab: exp_nlvr2}
\end{table}

\noindent{\textbf{Natural Language for Visual Reasoning (NLVR).}}
Following previous work~\cite{li2019visual}, we evaluate our CAVL pre-trained models on the NLVR$^2$ dataset for joint reasoning about natural language and images. 
In this task, we focus on predicting whether a natural language caption match with a pair of images. 
We report the comparison results with state-of-the-art methods in Table~\ref{tab: exp_nlvr2}.
As can be seen, our CAVL achieves the new state-of-the-art performance in terms of all experimental settings compared to previous methods. 
This demonstrates the effectiveness of the pair-wise contrastive loss incorporated in both visual and linguistic branches to mitigate the existing semantic confusion between vision and language at the pre-training stage.
The results also show the advantage of our CAVL in joint reasoning about language and vision.

\noindent{\textbf{Region-to-Phrase Grounding  (RPG).}}
In order to test the performance of our CAVL on RPG, we fine-tune our CAVL pre-trained models on the Flickr30K~\cite{yong2014from} Entities dataset consisting of 30k images and 250k annotations. 
Following BAN~\cite{kim2018bilinear} and VisualBERT~\cite{li2019visual}, we utilize image features from a pre-trained Faster R-CNN. For fine-tuning on Flickr30K Entities dataset, a self-attention block is applied to generate the average attention weights to predict the alignment between bounding boxes and phrases. Therefore, the box with the most attention from the phrase is the model prediction for a phrase to be grounded.

In this task, we need to choose the right bounding regions in an image that spans from a sentence belong to.
Table~\ref{tab: exp_flickr} reports the comparison results with existing methods.
We can observe that our CAVL outperforms previous multi-modal methods by a large margin under the same pre-trained setting, which demonstrates the effectiveness of our CAVL in visual-linguistic grounding tasks.

\begin{table}[t]
	%\normalem
	\renewcommand\tabcolsep{6.0pt}
	\centering
	\scalebox{0.8}{
	\begin{tabular}{lllllll}
		\toprule
		\multicolumn{1}{l}{\multirow{2}{*}{Model}} & \multicolumn{2}{c}{R@1} & 
		\multicolumn{2}{c}{R@5} & \multicolumn{2}{c}{R@10}                     \\
		& Dev & Test & Dev & Test & Dev & Test \\ 	\midrule
		BAN~\cite{kim2018bilinear} & - & 69.70 & - & 84.20 & - & 86.40\\
		VisualBERT~\cite{li2019visual}  &70.40 & 71.33 & 84.49 & 84.98 & 86.31 & 86.51 \\
		MDETR~\cite{aishwarya2021mdetr} & 78.90 & - & 88.80 & - & 90.80 & - \\
		CAVL (ours) & \textbf{80.35} & \textbf{81.76} & \textbf{90.63} & \textbf{91.12} & \textbf{93.16} & \textbf{94.21}  \\
		\bottomrule
		\end{tabular}
	}
	\caption{Comparison results on the Flickr30K Entities dataset.}
	\label{tab: exp_flickr}
\end{table}
% \vspace{-0.5em}
\begin{figure*}[!htb]

\setlength{\belowcaptionskip}{-0.5em}
	\centering
% 		\fbox{\rule{0pt}{2in} \rule{0.8\linewidth}{0pt}}
		\includegraphics[width=0.85\textwidth]{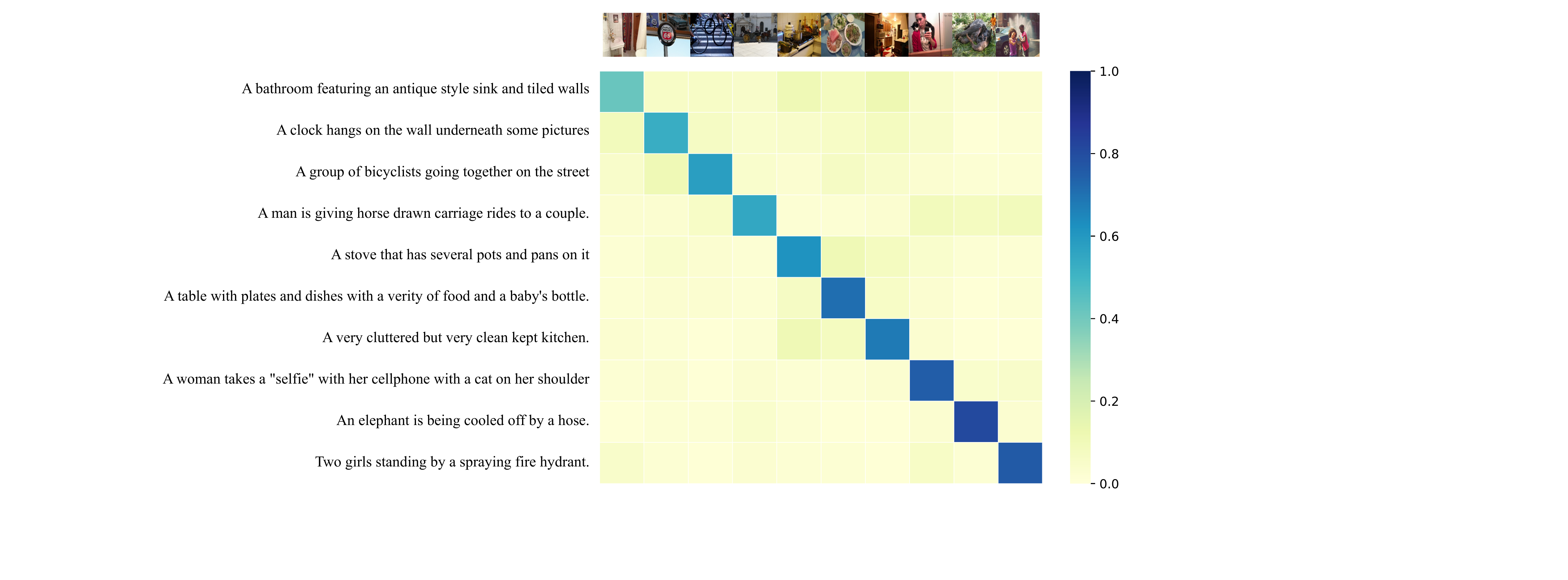}
% 		\vspace{-1em}
		 \caption{Heatmap visualization of cosine similarities between image and text pre-trained representations. Alignments across image and text pairs are learned during pre-training. 
    	}
	\label{fig: exp_vis}
\end{figure*}

\noindent{\textbf{Text-to-Image Retrieval (TIR).}}
Following previous work~\cite{lu2019vilbert,lu2021visualsparta}, we evaluate our CAVL on the Flickr30K~\cite{yong2014from} dataset for text-to-image retrieval and use recall rate (Recall@1, 5, 10) as the evaluation metrics. 
In this task, we need to match an image to a caption that describes its content.
Specifically, we calculate the Averaged Pair-wise Similarity (APS) score for each caption and one image.
In order to choose the right caption-image pair, We fine-tune the model with a
cross-entropy loss.
At inference stage in the test set, we sort each caption-image pair according to the alignment score.
The comparison results are reported in Table~\ref{tab: exp_retrieval}.
As can be seen, our CAVL achieves the best performance compared to previous baselines.
With the help of alleviating the semantic confusion, we also achieve comparable performance to ERNIE-ViL~\cite{Yu2021ERNIEViL}.
This demonstrates the effectiveness of our approach in Text-to-Image Retrieval.

\begin{table}[!htb]
	%\normalem
	\renewcommand\tabcolsep{6.0pt}
	\centering
	\scalebox{0.8}{
	\begin{tabular}{lllllll}
		\toprule
		\multicolumn{1}{l}{\multirow{2}{*}{Model}} & \multicolumn{3}{c}{TIR} & 
		\multicolumn{3}{c}{ZS-TIR}                     \\
		& R@1 & R@5 & R@10 & R@1 & R@5 & R@10 \\ 	\midrule
		
	SCAN~\cite{lee2018stacked} & 48.60 & 77.70 & 85.20 & - & - & - \\
	PFAN~\cite{wang2019position} & 50.40 & 78.70 & 86.10 & - & - & - \\
    CAMP~\cite{wang2019camp} & 51.50 & 77.10 & 85.30 & - & - & - \\
	CVSE~\cite{wang2020consensus} & 52.90 & 80.40 & 87.80 & - & - & -\\
	Oscar~\cite{li2020oscar} & 54.78 & 81.05 & - & - & - & - \\
	VisualSparta~\cite{lu2021visualsparta} & 57.10 & 82.60 & 88.20 & - & - & - \\
	VinVL~\cite{zhang2021vinvl} & 58.18 & 83.36 & - & - & - & - \\
		ViLBERT~\cite{lu2019vilbert} & 58.20 & 84.90 & 91.52 & 31.86 & 61.12 & 72.80 \\
		 SGRAF~\cite{Diao2021SGRAF} & 58.50 & 83.00 & 88.80 & - & - & - \\
		 Unicoder-VL~\cite{li2020unicodervl} & 71.50 & 91.20 & 95.20 & 48.40 & 76.00 & 85.20 \\
		 UNITER~\cite{uniter} & 72.52 & 92.36 & 96.08 & 66.16 & 88.40 & 92.94 \\
% 		 ERNIE-ViL~\cite{Yu2021ERNIEViL} & 74.44 & 92.72 & 95.94 & - & - & - \\ 
		CAVL (ours) & \textbf{74.06} & \textbf{92.56} & \textbf{96.15} & \textbf{67.07} & \textbf{88.92} & \textbf{93.36} \\
		\bottomrule
		\end{tabular}
	}
	\caption{Comparison results of Text-to-Image Retrieval (TIR) and Zero-shot Text-to-Image Retrieval (ZS-TIR) on the Flickr30K.}
	\label{tab: exp_retrieval}
	\vspace{-0.5em}
\end{table}

\noindent{\textbf{Zero-shot Text-to-Image Retrieval (ZS-TIR).}}
Furthermore, we follow previous work~\cite{lu2019vilbert}, and evaluate our CAVL on the Flickr30K~\cite{yong2014from} dataset for zero-shot text-to-image retrieval.
In this setting, we directly use the pre-trained weights without fine-tuning to compute the Averaged Pair-wise Similarity (APS) score for zero-shot retrieval.
We report the results in Table~\ref{tab: exp_retrieval}.
We also achieve a new state-of-the-art performance in terms of all metrics, which shows the effectiveness of our CAVL.

\subsection{Visualizations}

In this part, we visualize the pre-trained image and text pairs to validate the effectiveness of our CAVL in mitigating the semantic confusion between those representations. 
Specifically, we calculate the pair-wise similarity across pre-trained image and text pairs and report them in Figure~\ref{fig: exp_vis}.
As can be seen, our CAVL can learn alignments between the whole sentence and each image during pre-training.

\begin{table*}[h]
	%\normalem
	\renewcommand\tabcolsep{6.0pt}
	\centering
	\scalebox{0.8}{
	\begin{tabular}{cccccccc}
		\toprule
	    MLM & NSP & PwCL & AF & batch size & test-dev ($\uparrow$) & test-std  ($\uparrow$) & APS ($\uparrow$) \\
		\midrule
		\checkmark & \checkmark & &  & 64 & 70.11$\pm$0.12 & 71.03$\pm$0.15 & 0.43$\pm$0.08\\
		\checkmark &  & \checkmark & & 64 & 70.82$\pm$0.13 & 71.56$\pm$0.15 &  0.52$\pm$0.06\\
		 & \checkmark & \checkmark & & 64 & 70.06$\pm$0.12 & 70.95$\pm$0.13 & 0.41$\pm$0.06\\
		\checkmark & \checkmark & \checkmark & &  64 & 71.73$\pm$0.15 & 72.08$\pm$0.17 & 0.68$\pm$0.04\\
		\checkmark & \checkmark & \checkmark & \checkmark &  64 & 71.68$\pm$0.14 & 72.02$\pm$0.16 & 0.67$\pm$0.05\\
		\hline
	    \checkmark & \checkmark & \checkmark & \checkmark & 128 & 72.18$\pm$0.13 & 72.32$\pm$0.16 & 0.72$\pm$0.04\\
	    \checkmark & \checkmark & \checkmark & \checkmark & 256 & 72.42$\pm$0.11 & 72.67$\pm$0.13 & 0.76$\pm$0.03\\
	    \checkmark & \checkmark & \checkmark &  \checkmark & 512 & 72.83$\pm$0.05  & \textbf{73.05$\pm$0.07} & \textbf{0.83$\pm$0.02}\\
	    \checkmark & \checkmark & \checkmark &  \checkmark & 1024 & \textbf{72.88$\pm$0.08} & 72.96$\pm$0.06 & 0.81$\pm$0.02\\
		\bottomrule
		\end{tabular}
	}
	\caption{Ablation study on pair-wise contrastive pre-training, adaptive fine-tuning and batch size. MLM, NSP, PwCL, and AF denote Masked Language Modeling, Next Sentence Prediction, Pair-wise Contrastive Loss and Adaptive Fine-tuning.}
	\label{tab: ab_cavl}
	
\end{table*}

\begin{table*}[!htb]
	%\normalem
	\renewcommand\tabcolsep{6.0pt}
	\centering
	\scalebox{0.8}{
	\begin{tabular}{lcccc}
		\toprule
		Model & Fine-tune Params (M)$\downarrow$ & test-dev $\uparrow$ & test-std $\uparrow$ & Fine-tune Cost (h) $\downarrow$ \\
		\midrule
		VisualBERT~\cite{li2019visual} &113.90 & 70.08 &71.00 & 24.5 \\
		VL-BERT$_{\text{base}}$~\cite{Su2020VL-BERT:}  &115.04 &71.16 &- & 25.6\\
		VL-BERT$_{\text{large}}$~\cite{Su2020VL-BERT:}  & 342.55 &71.79 & 72.22 & 70.2\\\hline
		CAVL (Adapter I)  & 98.40 & \textbf{72.83$\pm$0.05} & \textbf{73.05$\pm$0.06} & 10.0\\
		CAVL (Adapter II) & \textbf{46.70} & 72.67$\pm$0.08 & 72.86$\pm$0.11 & \textbf{6.1}  \\
		\bottomrule
		\end{tabular}
	}
	\caption{Ablation study on variants of adapters proposed in our CAVL.}
	\label{tab: exp_adapter}
	\vspace{-0.5em}
\end{table*}

\subsection{Ablation Study}\label{ab:adpater}
In this section, we perform extensive ablation studies on the effect of pair-wise contrastive pre-training, adaptive fine-tuning and batch size on the final performance of our CAVL, and the efficiency of the proposed adapters (Adapter I and II). 
Unless specified, we conduct all ablation studies on the VQA 2.0 dataset and report the mean and standard deviation of all results with $5$ random seeds.

\noindent\textbf{Effect of each module and batch size.}
In Table~\ref{tab: ab_cavl}, we explore the effect of each pre-training task proposed in our CAVL, which consists of Masked Language Modeling, Next Sentence Prediction, Pair-wise Contrastive Loss, and Adaptive Fine-tuning. 
We can observe that with the incorporation of the pair-wise contrastive loss, our CAVL achieves better performance than the baseline without the pair-wise contrastive loss between linguistic and visual embeddings.
This demonstrates the effectiveness of the pair-wise contrastive loss proposed in our CAVL.
Our CAVL with adaptive fine-tuning achieves comparable performance while achieving fewer training parameters and saving computation resources.

We also evaluate the effect of the batch size on the final performance of our CAVL pre-trained models in Table~\ref{tab: ab_cavl}.
As can be seen, our CAVL performs the best APS at the batch size of 512, which shows the importance of the choice of the batch size in the pair-wise contrastive loss.
Adding PwCL to baselines with the same batch size increases the accuracy from 70.11 to 71.73, where the improvement (1.62) is significant in the VQA task. Increasing the batch size from 64 to 1024 can boost the accuracy from 71.68 to 72.83 (1.15), but the improvement is smaller than that of PwCL.
This also conforms to the importance of larger batch size in vision-language pre-training by introducing more negative pairs. We also show the big improvements of our CAVL in VCR (3.45), NLVR2 (11.31), and RPG (10.43) in Table~\ref{tab: exp_vqa_vcr},~\ref{tab: exp_nlvr2} and~\ref{tab: exp_flickr}.

Furthermore, we compare the Averaged Pair-wise Similarity (APS) between embeddings of each text-image pair during pre-training in Table~\ref{tab: ab_cavl}.
Specifically, we calculate the pair-wise dot product between linguistic embeddings $\mathbf{E}_i^\prime$ and visual embeddings $\mathbf{F}_i^\prime$, \textit{i.e.}, $\frac{1}{B}\sum_{i=1}^B (\mathbf{E}_i^\prime \cdot \mathbf{F}_i^\prime)$.
We can observe that our CAVL with the added pair-wise contrastive loss help increase the APS between each text-image pair, which verifies the effectiveness of our pair-wise contrastive pre-training in mitigating the semantic confusion between visual and linguistic embeddings during the pre-training process.

% \vspace{-0.5em}

\noindent\textbf{Effect of adapters.}
As shown in Table \blue{\ref{tab: exp_adapter}}, we compare the performance of various models in terms of training parameters, fine-tuning costs, and accuracy on test-dev, test-std sets. 
We observe that all CAVL based models achieve better results than the baselines. 
This further demonstrates the effectiveness of contrastive visual and linguistic pre-training in eliminating the semantic confusion during the pre-training process.
The proposed CAVL jointly learns alignments between the whole sentence and each image in the same batch to improve the pre-trained model's generalizability.
In the meanwhile, our CAVL with both adapters with fewer training parameters achieves comparable performance to baselines with large fine-tuning parameters, which validates the efficiency of our proposed adapters on fine-tuning fewer parameters to save computation resources. 

We also compare our CAVL with two types of adapters with current multi-modal methods~\cite{li2019visual,Su2020VL-BERT:} in terms of the average fine-tuning cost to evaluate how much time our CAVL could save during the fine-tuning phase.
Typically, both CAVL variants (Adapter I /II) achieves better results than previous work on both test-dev and test-std settings, which reducing remarkable parameters and costs for fine-tuning. 
While Adapter I performs slightly better than Adapter II, Adapter II reduces the fine-tuning parameters and costs by a large margin, \textit{i.e.}, 59.40\% and 76.17\%.
This implies the efficiency of our CAVL in the fine-tuning stage to learn effective representations for visual-linguistic downstream tasks.

% \vspace{-1em}
\section{Conclusion}
In this work, we propose a simple but effective framework for learning contrastive and adaptive representations of vision and language, called CAVL, which involves contrastive pre-training and adaptive fine-tuning. 
The pair-wise contrastive loss is applied to mitigate the semantic confusion between language and vision during pre-training.
Furthermore, we successfully introduce two lightweight adapters to eliminate the need for much computation time at the fine-tuning stage.
Our CAVL achieves superior performance against baselines on six main vision-language downstream tasks.
We conduct extensive experiments and ablation studies to demonstrate the efficiency of contrastive pre-training and adaptive fine-tuning.

%%%%%%%%% REFERENCES
{\small
\bibliographystyle{ieee_fullname}
\bibliography{reference}

\begin{thebibliography}{10}\itemsep=-1pt

\bibitem{anderson2018butd}
Peter Anderson, Xiaodong He, Chris Buehler, Damien Teney, Mark Johnson, and
  Stephen Gouldand~Lei Zhang.
\newblock Bottom-up and top-down attention for image captioning and visual
  question answering.
\newblock In {\em Proceedings of IEEE Conference on Computer Vision and Pattern
  Recognition}, page 6077–6086, 2018.

\bibitem{srikar2021docformer}
Srikar Appalaraju, Bhavan Jasani, Bhargava~Urala Kota, Yusheng Xie, and R.
  Manmatha.
\newblock Docformer: End-to-end transformer for document understanding.
\newblock {\em arXiv preprint arXiv:2106.11539}, 2021.

\bibitem{brown2020language}
Tom~B. Brown, Benjamin Mann, Nick Ryder, Melanie Subbiah, Jared Kaplan,
  Prafulla Dhariwal, Arvind Neelakantan, Pranav Shyam, Girish Sastry, Amanda
  Askell, Sandhini Agarwal, Ariel Herbert-Voss, Gretchen Krueger, Tom Henighan,
  Rewon Child, Aditya Ramesh, Daniel~M. Ziegler, Jeffrey Wu, Clemens Winter,
  Christopher Hesse, Mark Chen, Eric Sigler, Mateusz Litwin, Scott Gray,
  Benjamin Chess, Jack Clark, Christopher Berner, Sam McCandlish, Alec Radford,
  Ilya Sutskever, and Dario Amodei.
\newblock Language models are few-shot learners.
\newblock {\em arXiv preprint arXiv:2005.14165}, 2020.

\bibitem{chen2020simple}
Ting Chen, Simon Kornblith, Mohammad Norouzi, and Geoffrey Hinton.
\newblock A simple framework for contrastive learning of visual
  representations.
\newblock In {\em International Conference on Machine Learning}, pages
  1597--1607, 2020.

\bibitem{chen2020big}
Ting Chen, Simon Kornblith, Kevin Swersky, Mohammad Norouzi, and Geoffrey
  Hinton.
\newblock {Big Self-Supervised Models are Strong Semi-Supervised Learners}.
\newblock In {\em Advances in Neural Information Processing Systems}, pages
  22243--22255, 2020.

\bibitem{chen2020mocov2}
Xinlei Chen, Haoqi Fan, Ross Girshick, and Kaiming He.
\newblock Improved baselines with momentum contrastive learning.
\newblock {\em arXiv preprint arXiv:2003.04297}, 2020.

\bibitem{uniter}
Yen{-}Chun Chen, Linjie Li, Licheng Yu, Ahmed~El Kholy, Faisal Ahmed, Zhe Gan,
  Yu Cheng, and Jingjing Liu.
\newblock {UNITER:} learning universal image-text representations.
\newblock {\em arXiv preprint arXiv:1909.11740}, 2019.

\bibitem{devlin2018bert}
Jacob Devlin, Ming-Wei Chang, Kenton Lee, and Kristina Toutanova.
\newblock {BERT}: Pref-training of deep bidirectional transformers for language
  understanding.
\newblock {\em arXiv preprint arXiv:1810.04805}, 2018.

\bibitem{Diao2021SGRAF}
Haiwen Diao, Ying Zhang, Lin Ma, and Huchuan Lu.
\newblock Similarity reasoning and filtration for image-text matching.
\newblock In {\em The {AAAI} Conference on Artificial Intelligence}, 2021.

\bibitem{goyal2017vqa}
Yash Goyal, Tejas Khot, Douglas Summers-Stay, Dhruv Batra, and Devi Parikh.
\newblock Making the v in vqa matter: Elevating the role of image understanding
  in visual question answering.
\newblock In {\em Proceedings of IEEE Conference on Computer Vision and Pattern
  Recognition}, page 6904–6913, 2017.

\bibitem{he2019moco}
Kaiming He, Haoqi Fan, Yuxin Wu, Saining Xie, and Ross Girshick.
\newblock Momentum contrast for unsupervised visual representation learning.
\newblock In {\em Proceedings of IEEE Conference on Computer Vision and Pattern
  Recognition}, pages 9729--9738, 2020.

\bibitem{hong2021recurrent}
Yicong Hong, Qi Wu, Yuankai Qi, Cristian Rodriguez-Opazo, and Stephen Gould.
\newblock {VLN BERT:} a recurrent vision-and-language bert for navigation.
\newblock In {\em Proceedings of IEEE Conference on Computer Vision and Pattern
  Recognition}, 2021.

\bibitem{houlsby2019parameterefficient}
Neil Houlsby, Andrei Giurgiu, Stanislaw Jastrzebski, Bruna Morrone, Quentin de
  Laroussilhe, Andrea Gesmundo, Mona Attariyan, and Sylvain Gelly.
\newblock Parameter-efficient transfer learning for {NLP}.
\newblock In {\em International Conference on Machine Learning}, pages
  2790--2799, 2019.

\bibitem{huo2021wenlan}
Yuqi Huo, Manli Zhang, Guangzhen Liu, Haoyu Lu, Yizhao Gao, Guoxing Yang,
  Jingyuan Wen, Heng Zhang, Baogui Xu, Weihao Zheng, Zongzheng Xi, Yueqian
  Yang, Anwen Hu, Jinming Zhao, Ruichen Li, Yida Zhao, Liang Zhang, Yuqing
  Song, Xin Hong, Wanqing Cui, Danyang Hou, Yingyan Li, Junyi Li, Peiyu Liu,
  Zheng Gong, Chuhao Jin, Yuchong Sun, Shizhe Chen, Zhiwu Lu, Zhicheng Dou, Qin
  Jin, Yanyan Lan, Wayne~Xin Zhao, Ruihua Song, and Ji-Rong Wen.
\newblock {WenLan}: Bridging vision and language by large-scale multi-modal
  pre-training.
\newblock {\em arXiv preprint arXiv:2103.06561}, 2021.

\bibitem{Jia2021ALiGN}
Chao Jia, Yinfei Yang, Ye Xia, Yi-Ting Chen, Zarana Parekh, Hieu Pham, Quoc~V.
  Le, Yun-Hsuan Sung, Zhen Li, and Tom Duerig.
\newblock Scaling up visual and vision-language representation learning with
  noisy text supervision.
\newblock In {\em International Conference on Machine Learning}, 2021.

\bibitem{aishwarya2021mdetr}
Aishwarya Kamath, Mannat Singh, Yann LeCun, Ishan Misra, Gabriel Synnaeve, and
  Nicolas Carion.
\newblock {MDETR} - modulated detection for end-to-end multi-modal
  understanding.
\newblock {\em arXiv preprint arXiv:2104.12763}, 2021.

\bibitem{khosla2020sup}
Prannay Khosla, Piotr Teterwak, Chen Wang, Aaron Sarna, Yonglong Tian, Phillip
  Isola, Aaron Maschinot, Ce Liu, and Dilip Krishnan.
\newblock Supervised contrastive learning.
\newblock In {\em Advances in Neural Information Processing Systems}, pages
  18661--18673, 2020.

\bibitem{kim2018bilinear}
Jin-Hwa Kim, Jaehyun Jun, and Byoung-Tak Zhang.
\newblock Bilinear attention networks.
\newblock In {\em Advances in Neural Information Processing Systems}, page
  1571–1581, 2018.

\bibitem{kingma2014adam}
Diederik~P Kingma and Jimmy Ba.
\newblock Adam: A method for stochastic optimization.
\newblock {\em arXiv preprint arXiv:1412.6980}, 2014.

\bibitem{krishna2017visualgenome}
Ranjay Krishna, Yuke Zhu, Oliver Groth, Justin Johnson, Kenji Hata, Joshua
  Kravitz, Stephanie Chen, Yannis Kalantidis, Li-Jia Li, David~A. Shamma,
  Michael~S. Bernstein, and Fei-Fei Li.
\newblock {Visual Genome:} connecting language and vision using crowdsourced
  dense image annotations.
\newblock {\em International Journal of Computer Vision}, 123:32–73, 2017.

\bibitem{lee2018stacked}
Kuang-Huei Lee, Xi Chen, Gang Hua, Houdong Hu, and Xiaodong He.
\newblock Stacked cross attention for image-text matching.
\newblock pages 201--216, 2018.

\bibitem{lei2021less}
Jie Lei, Linjie Li, Luowei Zhou, Zhe Gan, Tamara~L. Berg, Mohit Bansal, and
  Jingjing Liu.
\newblock {Less is More:} clipbert for video-and-language learning via sparse
  sampling.
\newblock In {\em Proceedings of IEEE/CVF Conference on Computer Vision and
  Pattern Recognition}, 2021.

\bibitem{li2020unicodervl}
Gen Li, Nan Duan, Yuejian Fang, Ming Gong, and Daxin Jiang.
\newblock Unicoder-vl: {A} universal encoder for vision and language by
  cross-modal pre-training.
\newblock In {\em The Thirty-Fourth {AAAI} Conference on Artificial
  Intelligence}, pages 11336--11344, 2020.

\bibitem{li2019visual}
Liunian~Harold Li, Mark Yatskar, Da Yin, Cho{-}Jui Hsieh, and Kai{-}Wei Chang.
\newblock Visualbert: {A} simple and performant baseline for vision and
  language.
\newblock {\em arXiv preprint arXiv:1908.03557}, 2019.

\bibitem{li2020oscar}
Xiujun Li, Xi Yin, Chunyuan Li, Pengchuan Zhang, Xiaowei Hu, Lei Zhang, Lijuan
  Wang, Houdong Hu, Li Dong, Furu Wei, Yejin Choi, and Jianfeng Gao.
\newblock Oscar: Object-semantics aligned pre-training for vision-language
  tasks.
\newblock In {\em European Conference on Computer Vision}, 2020.

\bibitem{coco}
Tsung{-}Yi Lin, Michael Maire, Serge~J. Belongie, Lubomir~D. Bourdev, Ross~B.
  Girshick, James Hays, Pietro Perona, Deva Ramanan, Piotr Doll{\'{a}}r, and
  C.~Lawrence Zitnick.
\newblock Microsoft {COCO:} common objects in context.
\newblock {\em arXiv preprint arXiv:1405.0312}, 2014.

\bibitem{lu2019vilbert}
Jiasen Lu, Dhruv Batra, Devi Parikh, and Stefan Lee.
\newblock Vilbert: Pretraining task-agnostic visiolinguistic representations
  for vision-and-language tasks.
\newblock In {\em Advances in Neural Information Processing Systems}, pages
  13--23, 2019.

\bibitem{lu2021visualsparta}
Xiaopeng Lu, Tiancheng Zhao, and Kyusong Lee.
\newblock Visualsparta: Sparse transformer fragment-level matching for
  large-scale text-to-image search.
\newblock In {\em Proceedings of the 58th Annual Meeting of the Association for
  Computational Linguistics}, 2021.

\bibitem{mo2021spcl}
Shentong Mo, Zhun Sun, and Chao Li.
\newblock Siamese prototypical contrastive learning.
\newblock In {\em Proceeedings of British Machine Vision Conference}, 2021.

\bibitem{mo2022pauc}
Shentong Mo, Zhun Sun, and Chao Li.
\newblock Rethinking prototypical contrastive learning through alignment,
  uniformity and correlation.
\newblock In {\em Proceeedings of British Machine Vision Conference}, 2022.

\bibitem{mo2023mcvt}
Shentong Mo, Zhun Sun, and Chao Li.
\newblock Multi-level contrastive learning for self-supervised vision
  transformers.
\newblock In {\em 2023 IEEE/CVF Winter Conference on Applications of Computer
  Vision (WACV)}, pages 2777--2786, 2023.

\bibitem{pfeiffer2020madx}
Jonas Pfeiffer, Ivan Vulić, Iryna Gurevych, and Sebastian Ruder.
\newblock Mad-x: An adapter-based framework for multi-task cross-lingual
  transfer, 2020.

\bibitem{imagebert}
Di Qi, Lin Su, Jia Song, Edward Cui, Taroon Bharti, and Arun Sacheti.
\newblock {ImageBERT:} cross-modal pre-training with large-scale
  weak-supervised image-text data.
\newblock {\em arXiv preprint arXiv:2001.07966}, 2020.

\bibitem{radford2021learning}
Alec Radford, Jong~Wook Kim, Chris Hallacy, Aditya Ramesh, Gabriel Goh,
  Sandhini Agarwal, Girish Sastry, Amanda Askell, Pamela Mishkin, Jack Clark,
  Gretchen Krueger, and Ilya Sutskever.
\newblock Learning transferable visual models from natural language
  supervision.
\newblock {\em arXiv preprint arXiv:2103.00020}, 2021.

\bibitem{radford2018gpt}
Alec Radford, Karthik Narasimhan, Tim Salimans, and Ilya Sutskever.
\newblock {Improving Language Understanding by Generative Pre-Training}.
\newblock 2019.

\bibitem{radford2019gpt2}
Alec Radford, Jeff Wu, Rewon Child, David Luan, Dario Amodei, and Ilya
  Sutskever.
\newblock {Language Models are Unsupervised Multitask Learners}.
\newblock 2019.

\bibitem{ren2015faster}
Shaoqing Ren, Kaiming He, Ross Girshick, and Jian Sun.
\newblock Faster r-cnn: Towards real-time object detection with region proposal
  networks.
\newblock In {\em Advances in Neural Information Processing Systems}, pages
  91--99, 2015.

\bibitem{shi2020contrastive}
Lei Shi, Kai Shuang, Shijie Geng, Peng Su, Zhengkai Jiang, Peng Gao, Zuohui Fu,
  Gerard de Melo, and Sen Su.
\newblock Contrastive visual-linguistic pretraining.
\newblock {\em arXiv preprint arXiv:2007.13135}, 2020.

\bibitem{kvlbert}
Dandan Song, Siyi Ma, Zhanchen Sun, Sicheng Yang, and Lejian Liao.
\newblock {KVL-BERT:} knowledge enhanced visual-and-linguistic {BERT} for
  visual commonsense reasoning.
\newblock {\em arXiv preprint arXiv:2012.07000}, 2020.

\bibitem{Su2020VL-BERT:}
Weijie Su, Xizhou Zhu, Yue Cao, Bin Li, Lewei Lu, Furu Wei, and Jifeng Dai.
\newblock {VL-BERT}: Pre-training of generic visual-linguistic representations.
\newblock In {\em International Conference on Learning Representations}, 2020.

\bibitem{sun2019videobert}
Chen Sun, Austin Myers, Carl Vondrick, Kevin Murphy, and Cordelia Schmid.
\newblock {VideoBERT}: A joint model for video and language representation
  learning.
\newblock {\em arXiv preprint arXiv:1904.01766}, 2019.

\bibitem{tan2019lxmert}
Hao Tan and Mohit Bansal.
\newblock {LXMERT}: Learning cross-modality encoder representations from
  transformers.
\newblock {\em arXiv preprint arXiv:1908.07490}, 2019.

\bibitem{ashish2017attention}
Ashish Vaswani, Noam Shazeer, Niki Parmar, Jakob Uszkoreit, Llion Jones,
  Aidan~N. Gomez, Lukasz Kaiser, and Illia Polosukhin.
\newblock Attention is all you need.
\newblock In {\em Advances in Neural Information Processing Systems}, page
  5998–6008, 2017.

\bibitem{wang2020consensus}
Haoran Wang, Ying Zhang, Zhong Ji, Yanwei Pang, and Lin Ma.
\newblock Consensus-aware visual-semantic embedding for image-text matching.
\newblock In {\em European Conference on Computer Vision}, pages 18--34, 2020.

\bibitem{wang2019position}
Yaxiong Wang, Hao Yang, Xueming Qian, Lin Ma, Jing Lu, Biao Li, and Xin Fan.
\newblock Position focused attention network for image-text matching.
\newblock In {\em Proceedings of the 28th International Joint Conference on
  Artificial Intelligence}, page 3792–3798, 2019.

\bibitem{wang2019camp}
Zihao Wang, Xihui Liu, Hongsheng Li, Lu Sheng, Junjie Yan, Xiaogang Wang, and
  Jing Shao.
\newblock Camp: Cross-modal adaptive message passing for text-image retrieval.
\newblock In {\em Proceedings of the IEEE International Conference on Computer
  Vision}, pages 5764--5773, 2019.

\bibitem{yong2014from}
Peter Young, Alice Lai, Micah Hodosh, and Julia Hockenmaier.
\newblock From image descriptions to visual denotations: New similarity metrics
  for semantic inference over event descriptions.
\newblock {\em Transactions of the Association for Computational Linguistics},
  2:67--78, 2014.

\bibitem{ernievil}
Fei Yu, Jiji Tang, Weichong Yin, Yu Sun, Hao Tian, Hua Wu, and Haifeng Wang.
\newblock Ernie-vil: Knowledge enhanced vision-language representations through
  scene graph.
\newblock {\em arXiv preprint arXiv:2006.16934}, 2020.

\bibitem{Yu2021ERNIEViL}
Fei Yu, Jiji Tang, Weichong Yin, Yu Sun, Hao Tian, Hua Wu, and Haifeng Wang.
\newblock Ernie-vil: Knowledge enhanced vision-language representations through
  scene graph.
\newblock In {\em The Thirty-Fourth {AAAI} Conference on Artificial
  Intelligence}, 2021.

\bibitem{zellers2019vcr}
Rowan Zellers, Yonatan Bisk, Ali Farhadi, , and Yejin Choi.
\newblock From recognition to cognition: Visual commonsense reasoning.
\newblock In {\em Proceedings of IEEE Conference on Computer Vision and Pattern
  Recognition}, page 6720–6731, 2019.

\bibitem{zhang2021vinvl}
Pengchuan Zhang, Xiujun Li, Xiaowei Hu, Jianwei Yang, Lei Zhang, Lijuan Wang,
  Yejin Choi, and Jianfeng Gao.
\newblock Vinvl: Revisiting visual representations in vision-language models.
\newblock In {\em Proceedings of the IEEE/CVF Conference on Computer Vision and
  Pattern Recognition}, pages 5579--5588, 2021.

\bibitem{zhang2020contrastive}
Yuhao Zhang, Hang Jiang, Yasuhide Miura, Christopher~D Manning, and Curtis
  Langlotz.
\newblock Contrastive learning of medical visual representations from paired
  images and text.
\newblock {\em arXiv preprint arXiv:2010.00747}, 2020.

\end{thebibliography}
}

\end{document}